\documentclass[%
 reprint,
 amsmath,amssymb,
 aps,
 prl,
]{revtex4-2}

\usepackage{graphicx}% Include figure files
\usepackage{dcolumn}% Align table columns on decimal point
\usepackage{bm}% bold math
\begin{document}
\title{Controlling nonlinear dynamical systems into arbitrary states using machine learning} 

\author{Alexander Haluszczynski}
\email{alexander.haluszczynski@gmail.com}
\affiliation{Ludwig-Maximilians-Universit\"at, Department of Physics, Schellingstra{\ss}e 4, 80799 Munich, Germany}
\affiliation{Allianz Global Investors, risklab, Seidlstra{\ss}e 24, 80335, Munich, Germany}
\author{Christoph R{\"a}th}
\affiliation{Institut f{\"u}r Materialphysik im Weltraum, Deutsches Zentrum f{\"u}r Luft- und Raumfahrt,
M{\"u}nchner Str. 20, 82234 Wessling, Germany}

\begin{abstract}
%The ability to control dynamical system has important %application in a variety of fields ranging from rocket %engineering to medical technology. 

We propose a novel and fully data driven control scheme which relies on machine learning (ML). Exploiting recently developed ML-based prediction capabilities of complex systems we demonstrate that nonlinear systems can be forced to stay in arbitrary dynamical target states coming from any initial state.     
We outline our approach using the examples of the Lorenz and the R\"ossler system and show how these systems can very accurately be brought not only to periodic but also to e.g. intermittent and different chaotic behavior.  
Having this highly flexible control scheme with little demands on the amount of required data on hand, we briefly discuss possible applications that range from engineering to medicine.  

%Classical approaches rely on the exact knowledge of the %underlying dynamics and thus limit the practical %applicability. Furthermore, the state of the system can %typically only be changed from a rather complex one such %as chaotic dynamics to more simple states such as %periodic orbits or fixed points. We show that by using a %machine learning based control mechanism, we can force %the system into a variety of dynamical states. 

\end{abstract}

\maketitle

The possibility to control nonlinear chaotic systems into stable states has been a remarkable discovery \cite{ott1990controlling,shinbrot1993using}. Based on the knowledge of the underlying equations, one can force the system from a chaotic state into a fixed point or periodic orbit by applying an external force. This can be achieved based on the pioneering approaches by Ott et. al \cite{ott1990controlling} or Pyragas \cite{pyragas1992continuous}. In the former, a parameter of the system is slightly changed when it is close to an unstable period orbit in phase space, while the latter continuously applies a force based on time delayed feedback. There have been many extensions of those basic approaches (see e.g. \cite{boccaletti2000control} and references
therein) including "anti-control" schemes \cite{schiff1994controlling}, that break up periodic or synchronized
motion. But all of them do not allow to control the system into well-specified, yet more complex target states such as chaotic or intermittent behavior.
Further, these methods either require exact knowledge about the system, i.e. the underlying equations of motion, or rely on phase space techniques for which very long time series are necessary.\\ 
In recent years, tremendous progress has been made in the prediction of nonlinear dynamical systems by means of machine learning (ML). It has been demonstrated that not only exact short term predictions over several Lyapunov times become possible, but also the long term behavior of the system (its "climate") can be reproduced with unexpected accuracy \cite{chattopadhyay2020data,vlachas2020backpropagation,sangiorgio2020robustness,herteux2020breaking,haluszczynski2019good,griffith2019forecasting,lu2018attractor} -- even for very high-dimensional systems \cite{pathak2018model,zimmermann2018observing}. While several ML techniques have successfully been applied to time series prediction, reservoir computing (RC) \cite{maass2002real,jaeger2004harnessing} can be considered as the so far best approach as it combines often superior performance with intrinsic advantages like smaller network size,  higher robustness, fast and comparably transparent learning \cite{bompas2020accuracy} and the prospect of highly efficient hardware realizations \cite{marcucci2020theory,tanaka2019recent,carroll2021adding}.\\      
Combining now ML-based predictions of complex systems with manipulation steps, we propose in this {\it Letter} a novel, fully data-driven approach for controlling nonlinear dynamical systems. In contrast to previous methods, this allows to obtain a variety of target states including periodic, intermittent and chaotic ones. Furthermore, we do not require the knowledge of the underlying equations. Instead, it is sufficient to record some history of the system that allows the machine learning method to be sufficiently trained. As previously outlined \cite{pathak2017using}, an adequate learning requires orders of magnitude less data than phase space methods.\\
In this work, the prediction is solely performed with RC \cite{lukovsevivcius2009reservoir}, where the input data $\textbf{u}(t)$ is mapped into a higher dimensional state $\textbf{r}(t)$ through the dynamics of a fixed random reservoir $\textbf{A}$. The reservoir state $\textbf{r}(t)$ is updated according to $\textbf{r}(t+ \Delta{t}) = \alpha \textbf{r}(t)  +  tanh(\textbf{A}\textbf{r}(t) + \textbf{W}_{in} \textbf{u}(t))$, where $\textbf{W}_{in}$ denotes the fixed input mapping function. Output $\textbf{v}(t + \Delta{t})$ is created by mapping back $\textbf{r}(t)$ using a linear output function $\textbf{W}_{out}$ such that $\textbf{v}(t) = \textbf{W}_{out}(\mathbf{\tilde{r}}(t),  \textbf{P}) = \textbf{P}\mathbf{\tilde{r}}(t)$, where $\mathbf{\tilde{r}} = \{\mathbf{r}, \mathbf{r}^{2}\}$. The matrix $\textbf{P}$ is determined in the training process. Replacing $\textbf{u}(t)$ in the $\textit{tanh}$ activation function above by $\textbf{P}\mathbf{\tilde{r}}(t)$ allows to create predictions of arbitrary length due to the recursive equation for the reservoir states $\textbf{r}(t)$. Further details are presented in the supplementary material.

%%% Fig %%%
\begin{figure*}[t!] 
  \begin{center}
    \includegraphics[width=1.0\linewidth]{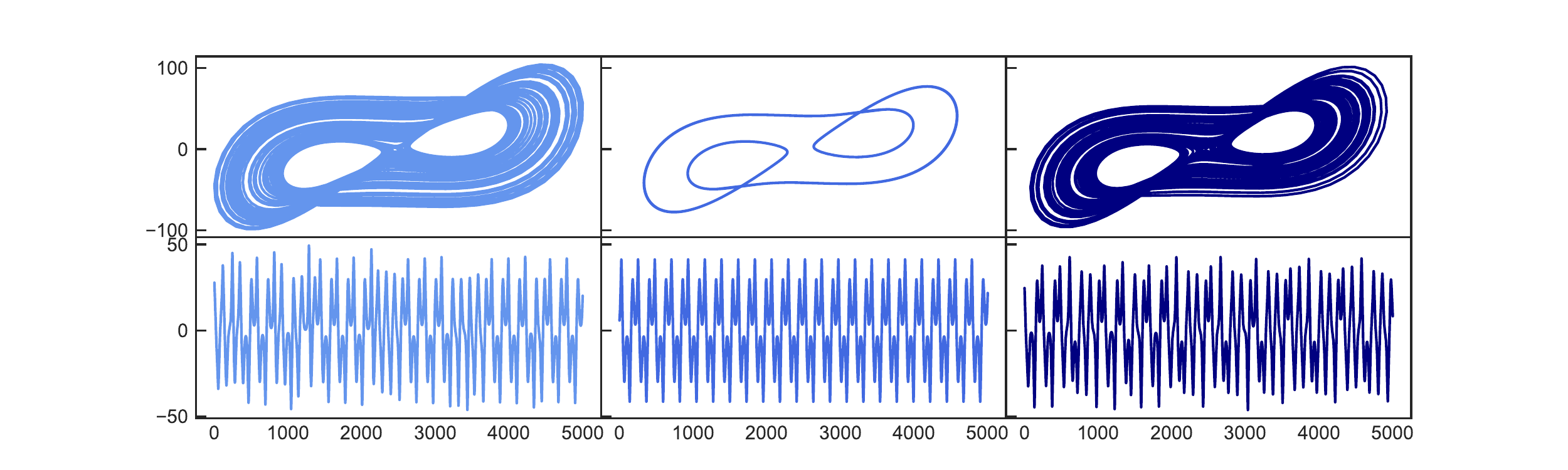}
    \caption{Periodic to chaotic control. Top: 2D attractor representation in the x-y plane. Bottom: X coordinate time series. Left plots show the original chaotic state which changes to a periodic state (middle) after tuning the order parameter. After applying the control mechanism, the system is forced into a chaotic state again (right).} 
    \label{fig:periodic2chaos}
  \end{center}
\end{figure*}
% Start main part here
The control of complex nonlinear dynamical system is studied on the example of the Lorenz system \cite{lorenz1963deterministic}, which is a model for atmospheric convection. Depending on the choice of parameters, the system exhibits e.g. periodic, intermittent or chaotic behavior. In contrast to classical approaches, our method relies on a good prediction of the system based on reservoir computing or other suitable machine learning techniques.
The setup for the control procedure is then the following: We simulate the Lorenz system for a certain parameter set $\bm{\pi}$ by integrating its differential equations $\dot{f}(\textbf{u},\bm{\pi})$ and train RC on it. Here, $\textbf{u} = (x,y,z)^{T}$ and the equations read
\begin{eqnarray}
\dot x = \sigma (y-x); \ \ \dot y = x (\rho-z)-y;\ \ \dot z = x y - \beta z \ ,
\label{eq:lorenz}
\end{eqnarray} 
while $\bm{\pi} \equiv (\sigma, \rho, \beta)$. Then the parameters are shifted to $\bm{\pi^{*}}$ such that the system behavior changes. In order to bring the system back to the desired original state, a correction force defined as $\textbf{F(t)}=K(\textbf{u}(t) - \textbf{v}(t))$ is applied. Now, $\textbf{u}(t)$  represents the coordinates of the system with changed parameters, whereas $\textbf{v}(t)$ denotes the theoretical coordinates of the predicted trajectory and $K$ scales the magnitude of the force and is set to $K=25$ for all examples. The coordinates $\textbf{u}(t+ \Delta{t})$ for the next time step are obtained by solving the equations of the Lorenz system including the force
\begin{eqnarray}
\textbf{u}(t+\Delta{t}) = \int_{t}^{t+\Delta{t}} (\dot{f}(\textbf{u}(\tilde{t}), \bm{\pi^{*}}) + \textbf{F}(\tilde{t})) d\tilde{t} ,
\end{eqnarray} 
however, still using the new parameters $\bm{\pi^{*}}$, which lead to an undesired state of the system if the control force $\textbf{F}(t)$ is not applied.
%%% Fig %%%
\begin{figure*}[t!] 
  \begin{center}
    \includegraphics[width=1.0\linewidth]{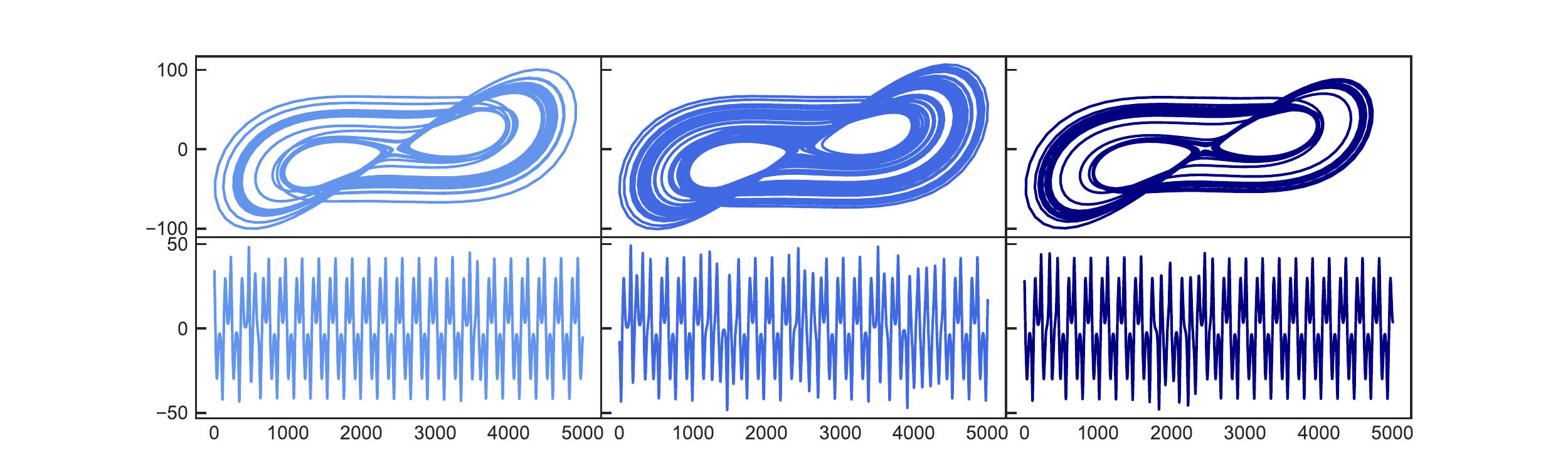}
    \caption{Chaotic to intermittent control. Top: 2D attractor representation in the x-y plane. Bottom: X coordinate time series. Left plots show the original intermittent state which changes to a chaotic state (middle) after tuning the order parameter. After applying the control mechanism, the system is forced into an intermittent state again (right).} 
    \label{fig:chaos2intermittent}
  \end{center}
\end{figure*}

Figure~\ref{fig:periodic2chaos} shows the results for the Lorenz system originally (left side) being in a chaotic regime ($\bm{\pi}=[\sigma=10.0,\rho=167.2,\beta=8/3]$), which then changes to periodic behavior (middle) after $\rho$ is changed to $\rho=166$. Then, the control mechanism is activated and the resulting attractor again resembles the original chaotic state (left). While 'chaotification' of periodic states has been achieved in the past, the resulting attractor generally did not correspond to a certain specified target state but just exhibited some chaotic behavior. Since we would like to not only rely on a visual assessment, we characterize the attractors using quantitative measures. First, we calculate the largest Lyapunov exponent, which quantifies the temporal complexity of the trajectory, where a positive value indicates chaotic behavior. Second, we use the correlation dimension to asses the structural complexity of the attractor. Based on the two measures, the dynamical state of the system can be sufficiently specified for our analysis. Both techniques are described in the supplementary material. Because a single example is not sufficiently meaningful, we perform our analysis statistically by evaluating 100 random realizations of the system at a time. The term 'random realization' refers to different random drawings of the reservoir $\textbf{A}$ and the input mapping $\textbf{W}_{in}$.
The first line in Table~\ref{tab:stats} shows the respective statistical results for the setup shown in Figure~\ref{fig:periodic2chaos}. The largest Lyapunov exponent of the original chaotic system $\lambda_{orig}= 0.851$ significantly reduces to $\lambda_{changed} = 0.080$ when the parameter change drives the system into a periodic state. After the control mechanism is switched on, the value for the resulting attractor moves back to $\lambda_{controlled} = 0.0841$ and thus is within one standard deviation from its original value. Same applies to the correlation dimension, which resembles its original value after control very well.
\begin{figure*}[t!] 
  \begin{center}
    \includegraphics[width=1.0\linewidth]{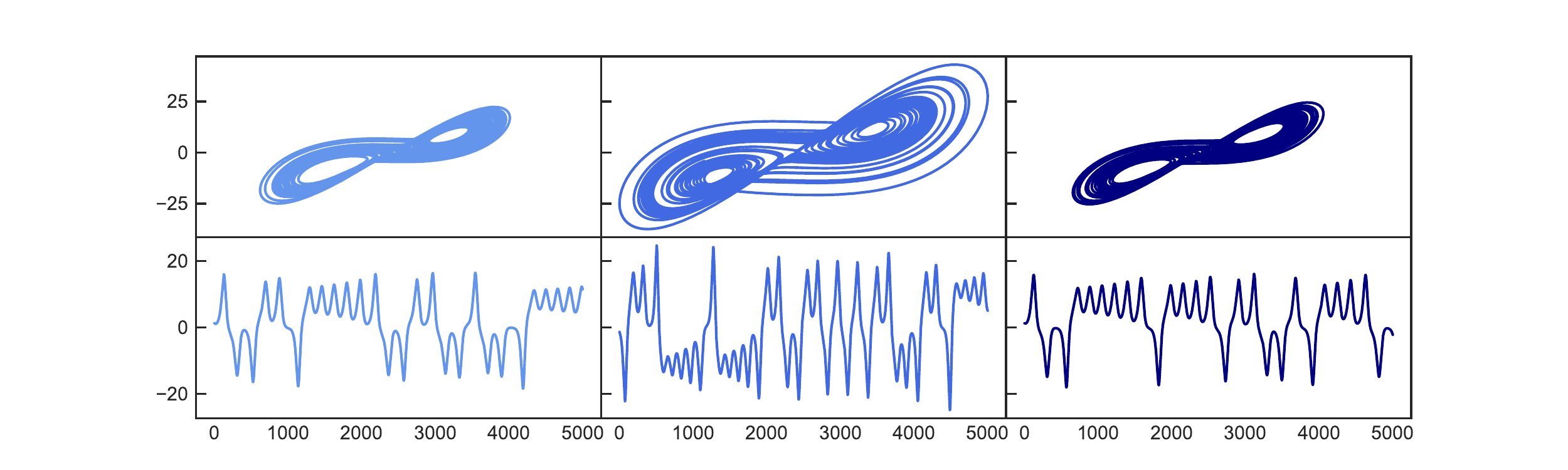}
    \caption{Chaotic to chaotic control. Top: 2D attractor representation in the x-y plane. Bottom: X coordinate time series. Left plots show the original chaotic state which changes to a different chaotic state (middle) after tuning the order parameter. After applying the control mechanism, the system is forced into the initial chaotic state again (right).} 
    \label{fig:chaos2chaos}
  \end{center}
\end{figure*}

Since there is a clear distinction between the chaotic- and the periodic state, with the latter being simple in terms of its dynamics, the next step is to control the system between more complex dynamics. Therefore, we start simulate the Lorenz system again with parameters $\bm{\pi}=[\sigma=10.0,\rho=166.15,\beta=8/3]$ that lead to intermittent behavior \cite{pomeau1980intermittent}. This is shown in Fig~\ref{fig:chaos2intermittent} on the left. Now $\rho$ is changed to $\rho=167.2$, which results in a chaotic state (middle plots). The control mechanism is turned on and the resulting state shows again the intermittent behavior (right plots) as in the initial state. This is particularly visible in the lower plots where only the X coordinate is shown. While the trajectory mostly follows a periodic path, it is interrupted by irregular burst that occur from time to time. It is remarkable that bursts do not seem to occur more often given the chaotic dynamics of the underlying equations and parameter setup. However, the control works so well that it exactly enforces the desired dynamics. This observation can again be confirmed by looking at the statistical results in Table~\ref{tab:stats}. 
\begin{figure}
  \begin{center}
    \includegraphics[width=1.0\linewidth]{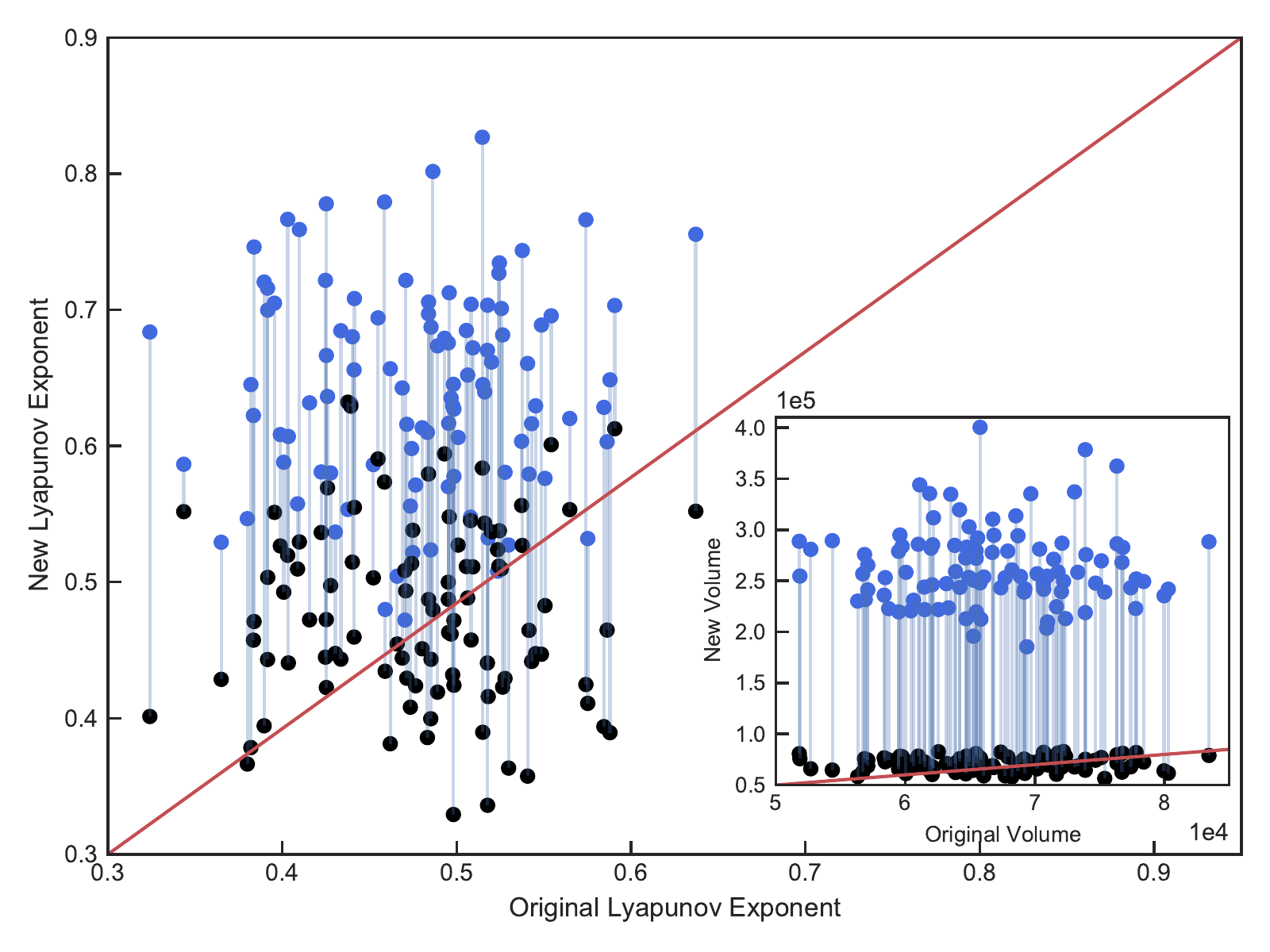}
    \caption{Chaotic to chaotic control ($\rho$ changed). Values on the x-axis denote the largest Lyapunov exponent $\lambda_{max}$ of the original system state before parameter change for $N=100$ random realizations. Y-axis reflects the values for $\lambda_{max}$ after parameters changed from $\rho=28$ to $\rho=50$. The blue dots correspond to the uncontrolled systems, while the black dots represent the controlled systems. Inlay plot shows the same for the volume of the attractor.} 
    \label{fig:chaos2chaosVol}
  \end{center}
\end{figure}
\newcommand{\size}[2]{{\fontsize{#1}{0}\selectfont#2}}
\begin{table*}
\caption{\label{tab:stats}Statistical simulation over $N=100$ random realizations of the systems evaluated in terms of the mean values of the largest Lyapunov exponent and the correlation dimension with corresponding standard deviations. The subscript \textit{orig} denotes the initial state of the system, while \textit{changed} refers to the new state after parameters changed and \textit{controlled} means the system controlled back into the original state. The description left to the arrow is the original state that also will be achieved again after controlling the system whereas the state written right to the arrow corresponds to the changed condition.}
\begin{ruledtabular}
\begin{tabular}{lcccccc}
 &\multicolumn{3}{c}{Largest Lyapunov Exponent $\lambda$}&\multicolumn{3}{c}{Correlation Dimension $\nu$}\\
 &$\lambda_{orig}$&$\lambda_{changed}$&$\lambda_{controlled}$&$\nu_{orig}$&$\nu_{changed}$&$\nu_{controlled}$\\ \hline
 $Periodic \rightarrow Chaotic$&0.851 \size{4}{$\pm 0.070$} &0.080 \size{4}{$\pm 0.075$}&0.841 \size{4}{$\pm 0.074$}&1.700\size{4}{$\pm 0.065$}&1.052 \size{4}{$\pm 0.071$}&1.700 \size{4}{$\pm 0.061$} \\
 $Chaotic \rightarrow Intermittent$&0.571 \size{4}{$\pm 0.096$}&0.853 \size{4}{$\pm 0.053$}&0.614 \size{4}{$\pm 0.101$}&1.321 \size{4}{$\pm 0.086$}&1.678 \size{4}{$\pm 0.055$}&1.351 \size{4}{$\pm 0.091$} \\ 
 $Chaotic_{B} \rightarrow Chaotic_{A}$&0.479 \size{4}{$\pm 0.060$}&0.643 \size{4}{$\pm 0.075$}&0.478 \size{4}{$\pm 0.067$}&1.941 \size{4}{$\pm 0.038$}&1.948 \size{4}{$\pm 0.047$}&1.933 \size{4}{$\pm 0.040$} \\
 $ Chaotic_{D} \rightarrow Chaotic_{C}$ &0.819 \size{4}{$\pm 0.092$}&0.884 \size{4}{$\pm 0.058$}&0.822 \size{4}{$\pm 0.052$}&1.855 \size{4}{$\pm 0.069$}&1.959 \size{4}{$\pm 0.037$}&1.866 \size{4}{$\pm 0.050$}\\
 \\
 $Periodic \leftarrow Chaotic$&-0.003 \size{4}{$\pm 0.012$}&0.844 \size{4}{$\pm 0.059$}&0.028 \size{4}{$\pm 0.110$}&1.001 \size{4}{$\pm 0.065$}&1.700 \size{4}{$\pm 0.071$}&1.001 \size{4}{$\pm 0.061$}\\
 $Chaotic \leftarrow Intermittent$&0.851 \size{4}{$\pm 0.070$}&0.550 \size{4}{$\pm 0.094$}&0.828 \size{4}{$\pm 0.067$}&1.700 \size{4}{$\pm 0.086$}&1.326 \size{4}{$\pm 0.055$}&1.698 \size{4}{$\pm 0.091$} \\ 
 $Chaotic_{B} \leftarrow Chaotic_{A}$&0.629 \size{4}{$\pm 0.069$}&0.446 \size{4}{$\pm 0.068$}&0.629 \size{4}{$\pm 0.066$}&1.948 \size{4}{$\pm 0.037$}&1.939 \size{4}{$\pm 0.049$}&1.956 \size{4}{$\pm 0.037$} \\
 $ Chaotic_{D} \leftarrow Chaotic_{C}$&0.881 \size{4}{$\pm 0.092$}&0.836 \size{4}{$\pm 0.058$}&0.880 \size{4}{$\pm 0.052$}&1.958 \size{4}{$\pm 0.069$}&1.864 \size{4}{$\pm 0.038$}&1.951 \size{4}{$\pm 0.050$}\\ 
\label{tab:stats}
\end{tabular}
\end{ruledtabular}
\end{table*}
Just like in the first two examples, it was not possible before to control a system from one chaotic state to another particular chaotic state. To do this, we start 
with the parameter set ($\bm{\pi}=[\sigma=10.0,\rho=28.0,\beta=8/3]$) leading to a chaotic attractor which we call $Chaotic_{A}$. When changing $\rho$ to $rho=50.0$ we obtain a different chaotic attractor $Chaotic_{B}$ This time we use a different range of values for $\rho$ compared to the previous examples in order to present a situation where not only the chaotic dynamics change, but also the size of the attractor significantly varies between the two states. The goal of the control procedure now is to not only force the dynamics of the system back to the behavior of the initial state $Chaotic_{A}$, but also to return the attractor to its original size. Figure~\ref{fig:chaos2chaos} shows that both goals succeed. This is also confirmed by the statistical results, indicating that the largest Lyapunov exponent of the controlled system is perfectly close to the one of the uncontrolled original state. For the correlation dimension, however, there are no significant deviations between the two chaotic states. To give a more striking illustration of the statistical analysis, we show the results for each of the 100 random realizations in Fig~\ref{fig:chaos2chaosVol}. The main plot scatters the largest Lyapunov exponents as measured for the original parameter set $\bm{\pi}$ agains those measured after the parameters have been changed to $\bm{\pi^{*}}$. While the blue dots represent the situation where the control mechanism is not active, the control has been switched on for the black dots. Furthermore, each pair of points is connected with a line that belongs to the same random realization. It is clearly visible that the control leads to a downwards shift of the cloud of points towards the diagonal, which is consistent to the respective average values of the largest Lyapunov exponent shown in Table~\ref{tab:stats}. In addition, the inlay plot shows the same logic but for the volume of the attractors being measured in terms of the smallest cuboid that covers the attractor. The control mechanism consistently works for every single realization and reduces the volume of the attractor back towards the initially desired state.
% Table with results from statistical analysis
% TODO: Make another table which summarizes parameters for all states mentioned

% TODO: Auf restlichen Teil von Tabelle 1 eingehen: Wir haben verschiedenste Zustände probiert und auch in die Gegenrichtung -> funtktioniert immer
The bottom half of Table~\ref{tab:stats} proves that our statements are also valid if one reverses the direction in the examples. For example, $Periodic \rightarrow Chaotic$ in the upper half of the table means, that an initially chaotic system changed into a periodic state and then gets controlled back into its initial chaotic state. In contrast,  $Periodic \leftarrow Chaotic$ in the lower half now means that the system initially is in the periodic state. It then shows chaotic behavior after the parameter change and finally is controlled back into the original periodic state - thus the opposite direction as above. It is evident that all examples also succeed in the opposite direction. This supports our claim that the prediction based control mechanism works for arbitrary states. Furthermore, we successfully tested our method also for the Roessler system and show the results in the supplementary material.

%TODO: Applications in engineering and physics. Need references and more concretely why our method outperforms existing ones
Our method has a wide range of potential applications in various areas. For example, in complex technical systems such as rocket engines it can be used to prevent the engine from critical combustion instabilities \cite{kabiraj2012route,nair2014intermittency}. This could be achieved by detecting them based on the reservoir computing predictions (or any other suitable ML technique) and subsequently controlling the system into a more stable state. Here, the control force can be applied to the engine via its pressure valves. Another example would be medical devices such as pacemakers. The heart of a healthy human does not beat in a purely periodic fashion but rather shows features being typical for chaotic systems like multifractality \cite{ivanov1999multifractality} that vary significantly among individuals. Our mechanism could therefore be used to develop personalized pacemakers that do not just stabilize the heartbeat to periodic behavior \cite{garfinkel1992controlling,hall1997dynamic,christini2001nonlinear}, but may rather adjust the heartbeat to the individual needs of the patients.\\ 
In conclusion, our machine learning enhanced method allows for an unprecedented flexible control of dynamical systems and has thus the potential to extend the range of applications of chaos inspired control schemes to a plethora of new real world problems. 

We would like to thank Y. Mabrouk, J. Herteux, S. Baur and J. Aumeier for helpful discussions.

\bibliographystyle{apsrev4-1} % Tell bibtex which bibliography style to use
\bibliography{bibliography}
\end{document}

%\preprint{}
\title{Supplemental materials to the manuscript}
\maketitle

\section{Reservoir Computing}
For the control mechanism to work it is crucial that good predictions of the system are available.  In the letter, we present the Lorenz system as an example of a complex nonlinear system which is controlled into different states. In order to obtain a good prediction of such a system, we use reservoir computing (RC) \cite{jaeger2001echo, maass2002real, jaeger2004harnessing}. RC is an artificial recurrent neural network based approach, which relies on a static internal network called $reservoir$. Static means that the nodes and edges are kept fixed once the network has been initially constructed. This property makes RC computationally very efficient, as only its linear output layer is being optimized in the training process. Thus, even high model dimensionality is computationally feasible, which makes the model well suited for complex real-world applications.

The implementation is mainly based on the setup of our previous study \cite{haluszczynski2019good}. The reservoir $\textbf{A}$ is constructed as a sparse Erd{\"o}s-Renyi random network \cite{erdos1959random} with dimensionality $D_{r}$. We chose to connect $D_{r}=300$ nodes that are connected with a probability $p=0.02$, such that we get an average unweighted degree of $d=6$. Initially, the weights of the edges are determined by independently drawn and uniformly distributed random numbers within the interval $[-1,1]$. A system parameter of particular interest is the spectral radius $\rho$ of the reservoir $\textbf{A}$, which is defined as its largest absolute eigenvalue 
\begin{eqnarray}
\rho(\textbf{A}) =  max \{|\lambda_{1}|, ..., |\lambda_{D_{r}}|\} \ ,
\label{eq:spectralradius}
\end{eqnarray} 
and can be interpreted as the average degree of the network. Given a fixed size $D_{r}$ of the reservoir, the magnitude of $\rho$ then depends on the number of edges and their weights. Therefore, one can adjust it as follows:
\begin{eqnarray}
\textbf{A}^{*} =  \frac{\textbf{A}}{\rho(\textbf{A})} \rho^{*} \ .
\label{eq:spectralradiuschange}
\end{eqnarray} 
Here, $ \rho^{*}$ is the desired spectral radius. In order to feed the $D$ dimensional input data $\textbf{u}(t)$ into the reservoir $\textbf{A}$, we set up an  $D_{r} \times D$ input mapping matrix $\textbf{W}_{in}$, which defines how strongly each input dimension influences every single node. The entries of $\textbf{W}_{in}$ are chosen to be uniformly distributed random numbers within the interval $[-\omega,\omega]$. We will specify $\omega$ and $ \rho^{*}$ in a second. The dynamics of the network are represented by its $D_{r} \times 1$  dimensional scalar states $\textbf{r}(t)$. Being initially set to $r_{i}(t_{0})=0$ for all nodes, they evolve according to the recurrent equation
\begin{eqnarray}
\textbf{r}(t+ \Delta{t}) = \alpha \textbf{r}(t)  + (1 -\alpha) tanh(\textbf{A}\textbf{r}(t) + \textbf{W}_{in} \textbf{u}(t)) \ .
\label{eq:updating}
\end{eqnarray} 
In this study we set $\alpha = 0$, and thus do not mix the input function (argument of the $tanh$, which is a function of $\textbf{r}(t)$) with past reservoir states $\textbf{r}(t)$ directly. In order to achieve optimal predictions, the choice for the scaling $\omega$ of the input function and the desired spectral radius $ \rho^{*}$ of the reservoir is crucial. As shown in our previous study \cite{ haluszczynski2020reducing}, the values should be chosen such that the distribution of the arguments of the hyperbolic tangent activation function lies within the dynamical range of the hyperbolic tangent function. This depends of course also on the ranges of the input data itself. If  $\omega$ and $ \rho^{*}$ were too large, then most values would be in the saturation regime of the $tanh$ function. In contrast, too small choices would lead to an approximately linear behavior. However, the nonlinearity of the activation function is an important requirement for good predictions of complex nonlinear systems. In this study we did not optimize for both parameters, but heuristically set reasonable values based on our insights from \cite{ haluszczynski2020reducing}, which depend on the example and are summarized in Table~\ref{tab:params}. 

In order to get the $D$ dimensional output $\textbf{v}(t)$ from the reservoir states $\textbf{r}(t)$, an output function $\textbf{W}_{out}$ is used that linearly depends on some output mapping matrix $\textbf{P}$
\begin{eqnarray}
\textbf{v}(t) = \textbf{W}_{out}(\textbf{r}(t),  \textbf{P}) = \textbf{P} \mathbf{\tilde{r}}(t) \ ,
\label{eq:output}
\end{eqnarray} 
where $\mathbf{\tilde{r}}(t)$ is a function of $\textbf{r}(t)$. Often $\mathbf{\tilde{r}}(t)=\textbf{r}(t)$ is used, but this leads to severe problems due to the antisymmetry of the hyperbolic tangent as explained in \cite{herteux2020reservoir}. To break this symmetry, we choose $\mathbf{\tilde{r}} = \{\mathbf{r}, \mathbf{r}^{2}\}$. This means that we append the squared elements of the reservoir states $\mathbf{r}^{2} = \{r_{1}^{2},r_{2}^{2},...,r_{D_r }^{2}\}$. The output mapping matrix $\textbf{P}$ then contains $2D_r \times D $ degrees of freedom and determining its coefficients is called \textit{training}. This is done by acquiring a sufficient number of reservoir states $\textbf{r}(t_{w}...t_{w}+t_{T})$ and then choosing $\textbf{P}$ such that the output $\textbf{v}$ of the reservoir is as close as possible to the known real data $\textbf{v}(t_{w}...t_{w}+t_{T})$. For this we use Ridge regression, which minimizes
 \begin{eqnarray}
\sum^{}_{-T \leq t \leq 0} {\parallel  \textbf{W}_{out}(\mathbf{\tilde{r}}(t), \textbf{P}) - \textbf{v}_{R}(t) \parallel}^2 - \beta {\parallel \textbf{P} \parallel}^2 \ ,
\label{eq:minimizing}
\end{eqnarray} 
where $\beta$ is the regularization constant that prevents from overfitting by penalizing large values of the fitting parameters. The notation $\parallel \textbf{P} \parallel$ describes the sum of the square elements of the matrix $\textbf{P}$. This ridge regression problem can be solved in matrix form \cite{hoerl1970ridge} reading
\begin{eqnarray}
\textbf{P} = {(\mathbf{\tilde{r}}^{T} \mathbf{\tilde{r}} + \beta \mathds{1})}^{-1} \mathbf{\tilde{r}}^{T} \textbf{v}_{R} \ .
\label{eq:ridgematrix}
\end{eqnarray} 
The notion $\textbf{r}$ and $\textbf{v}_{R}$ without the time indexing $t$ denotes matrices, where the columns contain the vectors $\textbf{r}(t)$ and $\textbf{v}_{R}(t)$ respectively for every time step. Before recording the reservoir states $\textbf{r}$, the system runs through an initialization or washout phase for $t_{w} = 5000$ time steps in order to allow for sufficient synchronization with the dynamics of the input signal $\textbf{u}$. After this, the RC system is trained for $t_{T} = 5000$ time steps. 

\begin{figure}[b!] 
  \begin{center}
    \includegraphics[width=0.7\linewidth]{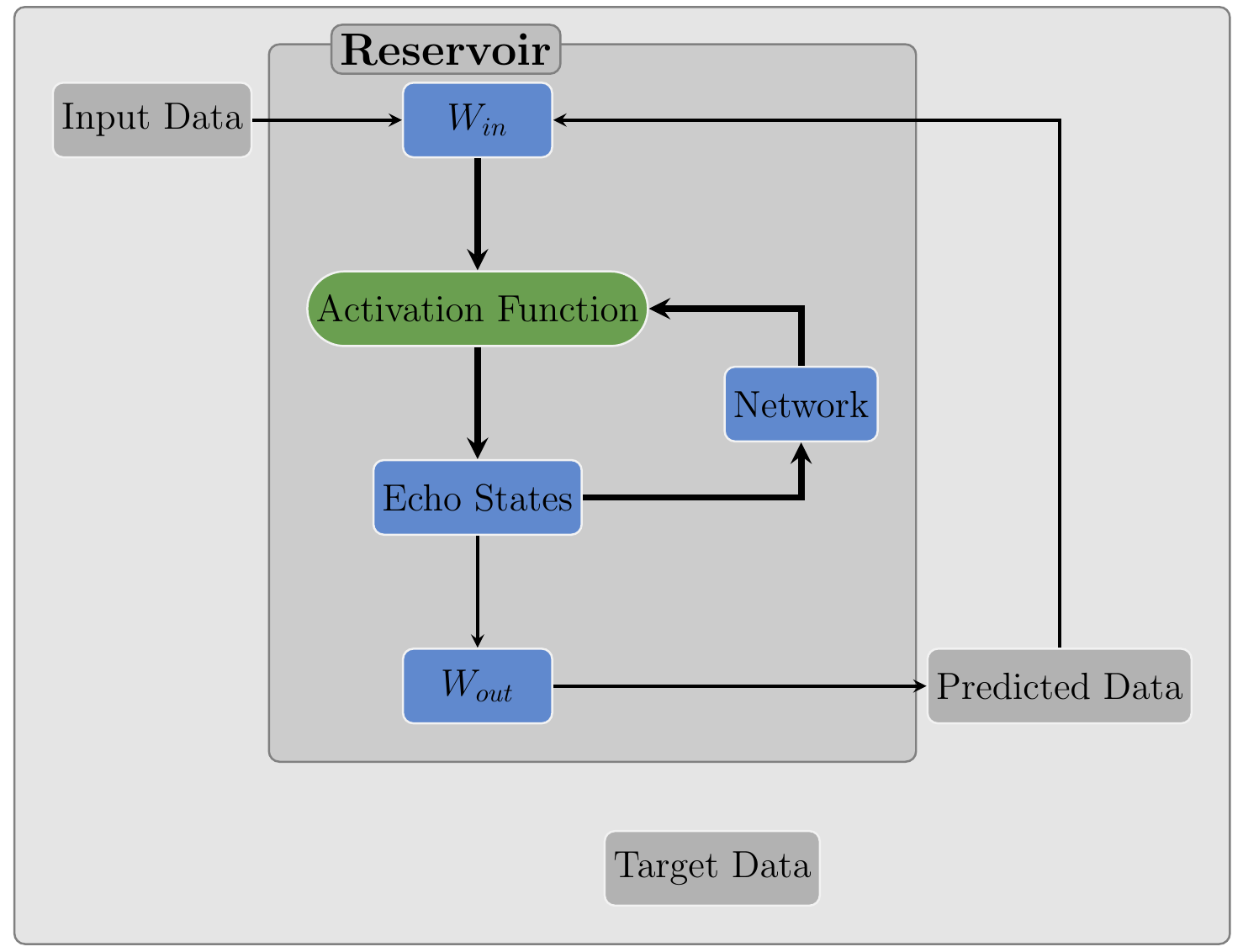}
    \caption{Schematic illustration of reservoir computing taken from \cite{ haluszczynski2020reducing}.}
    \label{fig:rescomp}
  \end{center}
\end{figure}

Once \textbf{P} has been trained, the predicted state $\textbf{v}(t)$ can be fed back in the activation function as input instead of the actual data $\textbf{u}(t)$ by combining Eq~\ref{eq:updating} and Eq~\ref{eq:output}. This allows to create predicted trajectories of arbitrary length due to the recursive equation for the reservoir states $\textbf{r}(t)$:
\begin{eqnarray}
%\begin{aligned}
\textbf{r}(t+ \Delta{t}) &= tanh(\textbf{A}\textbf{r}(t) + \textbf{W}_{in} \textbf{W}_{out}(\mathbf{\tilde{r}}(t),\textbf{P})) \\
                                  &= tanh(\textbf{A}\textbf{r}(t) + \textbf{W}_{in} \textbf{P}\mathbf{\tilde{r}}(t))  \ .
\label{eq:updatingprediction}
%\end{aligned}
\end{eqnarray} 
A schematic overview of the above described reservoir computing framework is given in Fig~\ref{fig:rescomp}.

\begin{table}
\caption{\label{tab:params} Summary of parameters of the reservoir computing predictions used in the examples shown in the letter}
\begin{ruledtabular}
\begin{tabular}{lccc}
 &$\omega$&$ \rho^{*}$ & $\beta$\\ \hline
 $Periodic \rightarrow Chaotic$&0.0084&0.0084&$6*10^{-11}$\\
 $Chaotic \rightarrow Intermittent$&0.0084&0.0084&$1*10^{-11}$ \\ 
 $Chaotic_{B} \rightarrow Chaotic_{A}$&0.0100&0.0100&$1*10^{-11}$ \\
 $ Chaotic_{D} \rightarrow Chaotic_{C}$ &0.0150&0.0150&$1*10^{-11}$\\
 \\
 $Periodic \leftarrow Chaotic$&0.0084&0.0084&$6*10^{-11}$ \\
 $Chaotic \leftarrow Intermittent$&0.0084&0.0084&$1*10^{-11}$ \\ 
 $Chaotic_{B} \leftarrow Chaotic_{A}$&0.0025&0.0025&$1*10^{-11}$ \\
 $ Chaotic_{D} \leftarrow Chaotic_{C}$&0.0120&0.0120&$1*10^{-11}$
\label{tab:params}
\end{tabular}
\end{ruledtabular}
\end{table}

% Fig %%%

\section{Lorenz System and Measures}

We use the Lorenz system \cite{lorenz1963deterministic} as an example for our control mechanism. It is a simple model for atmospheric convection and exhibits chaos or intermittent behavior for certain parameter ranges, while other parameter ranges lead to periodic trajectories. Thus it offers a wide variety of different dynamical states. The equations read
\begin{eqnarray}
\begin{aligned}
\dot x &= \sigma (y-x) \\
\dot y &= x (\rho-z)-y \\
\dot z &= x y - \beta z  \ .
\label{eq:lorenz}
\end{aligned}
\end{eqnarray} 
The standard parameter choices for a chaotic regime are $\sigma = 10 , \beta = 8/3$ and $\rho = 28$. We state in the letter which parameters we used in each example. The equations are solved using the 4th order Runge-Kutta method with a time resolution $\Delta t = 0.02$.  \\

To characterize the attractor and therefore its dynamical state we rely on quantitative measures. For this, we are looking at the long-term properties of the attractor rather than its short term trajectory. One important aspect of the long-term behavior is the structural complexity. This can be assessed by calculating the correlation dimension of the attractor, where we measure the dimensionality of the space populated by the trajectory \cite{grassberger2004measuring}. The correlation dimension is based on the correlation integral 
\begin{eqnarray}
\begin{aligned}
C(r) &= \lim\limits_{N \rightarrow \infty}{\frac{1}{N^2}\sum^{N}_{i,j=1}\theta(r- | \textbf{x}_{i} - \textbf{x}_{j}  |)}      \\
       &= \int_{0}^{r} d^3 r^{\prime} c(\textbf{r}^{\prime}) \ ,
\label{eq:corrintegral}
\end{aligned}
\end{eqnarray}
where $\theta$ is the Heaviside function and $c(\textbf{r}^{\prime})$ denotes the standard correlation function. The correlation integral represents the mean probability that two states in phase space are close to each other at different time steps. This is the case if the distance between the two states is less than the threshold distance $r$. The correlation dimension $\nu$ is then defined by the power-law relationship
\begin{eqnarray}
C(r) \propto r^{\nu} \ .
\label{eq:corrdim}
\end{eqnarray} 
For self-similar strange attractors, this relationship holds for a certain range of $r$, which therefore needs to be properly calibrated. As we are finally only interested in comparisons, precision with regards to absolute values is not essential here. We use the Grassberger Procaccia algorithm \cite{grassberger1983generalized} to calculate the correlation dimension.

In addition to the structural complexity, we are also interested in the temporal complexity of the attractor. This can be measured by its Lyapunov exponents $\lambda_{i}$. The Lyapunov exponents describe the average rate of divergence of nearby points in phase space, and thus measure sensitivity to initial conditions.   There is one exponent for each dimension in phase space. If the system exhibits at least one positive Lyapunov exponent, it is classified as chaotic. The magnitudes of $\lambda_{i}$ quantify the time scale on which the system becomes unpredictable \cite{wolf1985determining, shaw1981strange}. Since at least one positive exponent is the requirement for being classified as chaotic, it is sufficient for our analysis to calculate only the largest Lyapunov exponent $\lambda_{max}$
\begin{eqnarray}
d(t) =  C e^{\lambda_{max} t} \ .
\label{eq:lyapunov}
\end{eqnarray} 
This makes the task computationally much easier than determining the full Lyapunov spectrum. We use the Rosenstein algorithm \cite{rosenstein1993practical} to obtain it. In essence, we track the distance $d(t)$ of two initially nearby states in phase space. The constant $C$ normalizes the initial separation. As for the correlation dimension, we are interested in a relative comparison in order to characterizes states of the system rather than the exact absolute values. It is important to point out that both measures - the correlation dimension and the largest Lyapunov exponent - are calculated purely based on data and do not require any knowledge of the system. 

\section{Further Results}
\begin{figure}
  \begin{center}
    \includegraphics[width=1\linewidth]{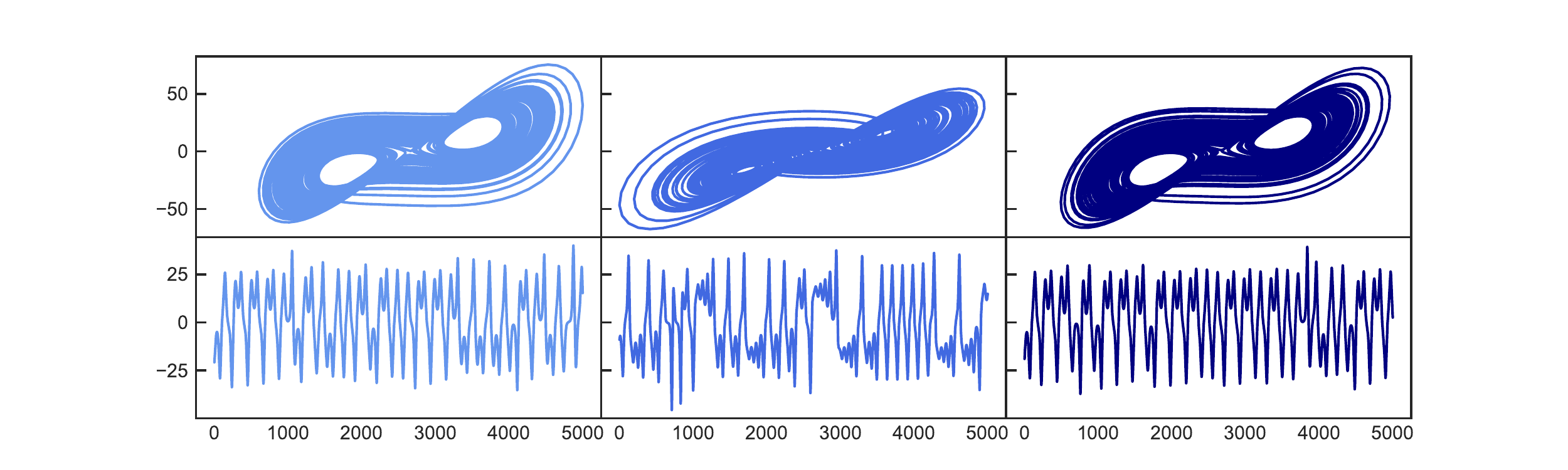}
    \caption{Chaotic to chaotic control. Top: 2D attractor representation in the x-y plane. Bottom: X coordinate time series. Left plots show the original chaotic state which changes to a different chaotic state (middle) after tuning the order parameter. After applying the control mechanism, the system is forced into the initial chaotic state again (right).} 
    \label{fig:chaos_sigma}
  \end{center}
\end{figure}

We presented three visual examples (plots of the attractors) for our control mechanism in the letter, while statistical results are provided for further examples. In particular, we tried some more parameter sets where a chaotic state is controlled to another different chaotic state. Figure~\ref{fig:chaos_sigma} shows an example where we did not vary $\rho$ as in the other examples but $\sigma$. This corresponds to the case $Chaotic_{D} \rightarrow Chaotic_{C}$ in the letter and the parameters used for the reservoir computing prediction can be found in Table~\ref{tab:params}. The initial parameters are $\bm{\pi}=[\sigma=10.0,\rho=102.0,\beta=8/3]$, which lead to a chaotic behavior of the Lorenz system. Those are then changed to $\bm{\pi^{*}}=[\sigma=20.0,\rho=102,\beta=8/3]$ leading to another chaotic state. The initial state is shown in the left plots of Fig~\ref{fig:chaos_sigma}, while the attractor based on the new parameters $\bm{\pi^{*}}$ is shown in the middle. Both attractors have approximately the same size but look different. This can particularly be observed when looking on the $x$ coordinates only (bottom plots). We can also quantify the differences based on the measures we introduced to characterize the temporal and structural complexity of an attractor: The largest Lyapunov exponent $\lambda_{max}$ and the correlation dimension $\nu$. While the initial state has the properties $[\lambda_{max}=0.90, \nu = 1.88]$, the second state is characterized by $[\lambda_{max}=0.84, \nu = 1.91]$ and therefore has a slightly lower largest Lyapunov exponent with the correlation dimension being quite similar. After the control mechanism is switched on, the attractor now (right plot) looks again like the initial attractor (left plot) although the simulation still runs with the changed parameters $\bm{\pi^{*}}$. However, the control force $F$, which is based on the reservoir computing prediction that has been trained on the initial parameter set $\bm{\pi}$, pushes the system successfully back into its desired initial state. The dynamical properties of the controlled attractor are $[\lambda_{max}=0.90, \nu = 1.87]$ and therefore the Lyapunov exponent takes on the same value as for the original attractor. The correlation dimension stays about the same, which can be a result of the fuzziness of the calculation method used.

\begin{figure}
  \begin{center}
    \includegraphics[width=1\linewidth]{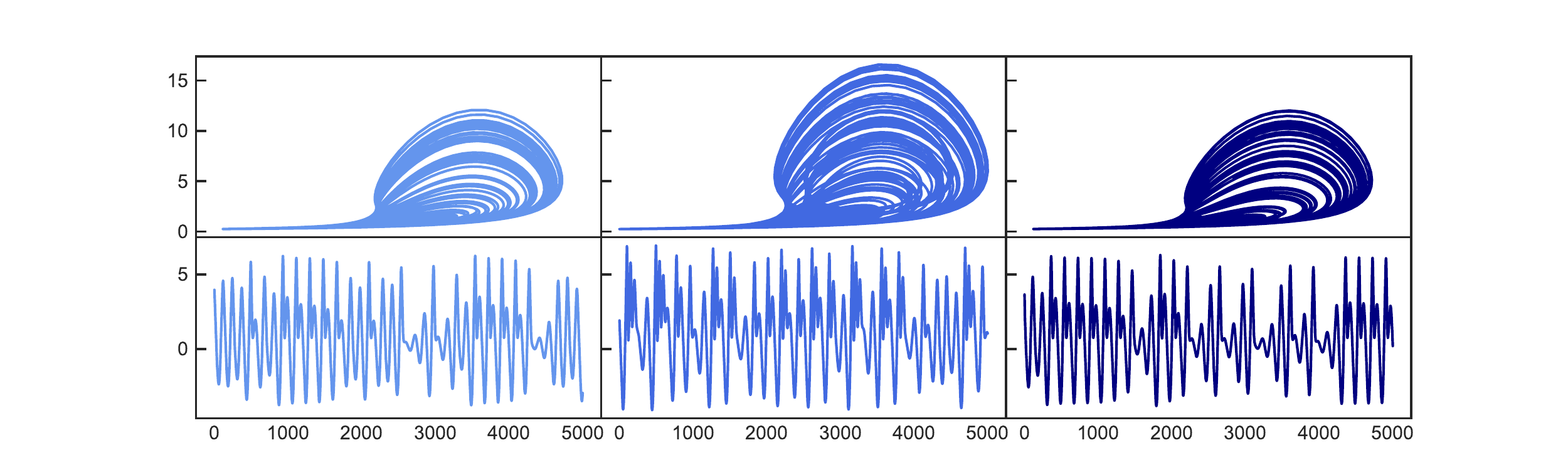}
    \caption{Chaotic to chaotic control for the Roessler system. Top: 2D attractor representation in the x-z plane. Bottom: X coordinate time series. Left plots show the original chaotic state which changes to a different chaotic state (middle) after tuning the order parameter. After applying the control mechanism, the system is forced into the initial chaotic state again (right).} 
    \label{fig:chaos_roessler}
  \end{center}
\end{figure}

In addition to the Lorenz system we also applied the method to another popular chaotic attractor: the Roessler system \cite{rossler1976equation}. The equations read
\begin{eqnarray}
\begin{aligned}
\dot x &= -(y+z) \\
\dot y &= x +ay \\
\dot z &= b+(x-c)z  \ .
\label{eq:roessler}
\end{aligned}
\end{eqnarray} 
and we use parameters $\bm{\pi}=[a=0.5,b=2.0,c=4.0]$ leading to a chaotic behavior. This serves as our initial state and the dynamics change to another chaotic state after the parameters are changed to $\bm{\pi^{*}}=[a=0.55,b=2.0,c=4.0]$. For the Roessler system we use a time resolution of $\Delta t = 0.5$ and $K=20$ for the control procedure, as well as the input scaling $\omega=0.8$ and the spectral radius $ \rho^{*}=0.4$ for the reservoir computing predictions. It can be seen in Fig~\ref{fig:chaos_roessler} that the control mechanism works. Again, the left plots represent the initial attractor resulting from the parameter set $\bm{\pi}$. Switching to $\bm{\pi^{*}}$ (middle plots) not only increases the size of the attractor in the x-z plane, but also significantly changes the pattern of the x-coordinate time series. Both, the appearance of the attractor and its x-coordinate pattern become similar to the initial attractor again after the control mechanism is active (right plots). The initial state with parameters $\bm{\pi}$ has properties $[\lambda_{max}=0.13, \nu = 1.59]$, which become $[\lambda_{max}=0.14, \nu = 1.75]$ after parameters have been changed to $\bm{\pi^{*}}$. Turning on the control mechanism leads to $[\lambda_{max}=0.12, \nu = 1.64$.

\bibliographystyle{apsrev4-1} % Tell bibtex which bibliography style to use
\bibliography{bibliographySup}